\documentclass[journal]{IEEEtran}

\usepackage{multirow}

\usepackage{cite}

\ifCLASSINFOpdf

\else

\fi

\newcommand{\e}[1]{\ensuremath{\times 10^{#1}}}

\ifCLASSOPTIONcompsoc
  \usepackage[caption=false,font=normalsize,labelfont=sf,textfont=sf]{subfig}
\else
  \usepackage[caption=false,font=footnotesize]{subfig}
\fi

\usepackage[T1]{fontenc}
\usepackage[utf8]{luainputenc}

\usepackage{afterpage}
\usepackage{float}
\usepackage{graphicx}
\usepackage{epstopdf}
\usepackage{epsfig}
\AppendGraphicsExtensions{.tif}

\usepackage{gensymb}
\usepackage{algorithm}
\usepackage{algorithmicx}
\usepackage{amsmath}
\usepackage{amssymb}

\usepackage{threeparttable}
\usepackage{array}
\usepackage{xcolor}
\usepackage{bm}
\usepackage{mathrsfs}
\usepackage{verbatim}
 \usepackage{diagbox} 
 
\usepackage{threeparttable}

\begin{document}

\title{Development of a Self-Calibrated Motion Capture System by Nonlinear Trilateration of Multiple Kinects v2}

\author{Bowen Yang, Haiwei Dong,~\IEEEmembership{Senior Member,~IEEE}, and Abdulmotaleb El Saddik,~\IEEEmembership{Fellow,~IEEE}
	
\thanks{B. Yang, H. Dong, and A. El Saddik are with Multimedia Computing Research Laboratory (MCRLab), School of Electrical Engineering and Computer Science, University of Ottawa, 800 King Edward, Ottawa, ON, K1N 6N5, Canada (e-mail: \{byang078; hdong; elsaddik\}@uottawa.ca).}
}

\maketitle


%

\begin{abstract}
	In this paper, a Kinect-based distributed and real-time motion capture system is developed. A trigonometric method is applied to calculate the relative position of Kinect v2 sensors with a calibration wand and register the sensors' positions automatically. By combining results from multiple sensors with a nonlinear least square method, the accuracy of the motion capture is optimized. Moreover, to exclude inaccurate results from sensors, a computational geometry is applied in the occlusion approach, which discovers occluded joint data. The synchronization approach is based on an NTP protocol that synchronizes the time between the clocks of a server and clients dynamically, ensuring that the proposed system is a real-time system. Experiments for validating the proposed system are conducted from the perspective of calibration, occlusion, accuracy, and efficiency. Furthermore, to demonstrate the practical performance of our system, a comparison of  previously developed motion capture systems (the linear trilateration approach and the geometric trilateration approach) with the benchmark OptiTrack system is conducted, therein showing that the accuracy of our proposed system is $38.3\%$ and $24.1\%$ better than the two aforementioned trilateration systems, respectively.
	
\end{abstract}


%
\IEEEpeerreviewmaketitle

\section{Introduction}

Motion capture technique is first proposed in late 1970s \cite{intro_mc_1}. It helps people to track human body's movement and apply the captured motion for further purposes \cite{intro_9}. Nowadays, it has been widely used in multiple fields: in the film industry, it is utilized to track actor's posture and facial expression to build the virtual character with animation or computer graphic \cite{inrto_mc_2}; in sport medicine, it is applied to analyze athletes' motion to promote their sport performance \cite{inrto_mc_3}; in entertainment, particularly in somatosensory game, the motion captured data is used to synchronize the motion of the virtual character with that of the video game player\cite{oh1997motion}.

Kinect is a series of motion sensing input devices developed by Microsoft for their XBox series game consoles and Windows PCs. Kinect recognizes and tracks the human body through the depth measurement to provide the motion capturing. It applies random decision forest \cite{inrtro_kinect_1,inrto_kinect_2} to build its human body tracking algorithm which maps depth image to the body parts and calculates the joints' positions to simulate the skeleton. It is developed on over 1 million training examples and can quickly track  3 people in a single depth image for 200 frames per second. Kinect v1 applies light coding\cite{Inrto_lightcoding} principle using infrared (IR) sensor to measure the depth data. It projects IR pulse constituted with numerous dots spaced in equal distance to the surface of the environment, and measures the distance and the shape of the surface with the distance of dots' spacing perceives. Kinect v2 applies the time-of-flight principle which combines the IR projector with a monochrome CMOS sensor. It resolves distance based on the known speed providing a better performance compared with Kinect v1. The accuracy of Kinect v2 depth measurement is not affected by the density of the dots in IR pulse which decreases with the increased distance, and thus it has a larger measurement range compared with that of  Kinect v1. Moreover, light coding has a more stringent requirement on setup conditions, such as the light intensity and the light absorption of the surface. The Kinect v2 also updates a more powerful chipset which supports a higher data rate of framing. It can transfer the depth measurement as $512 \times 424$ depth map stream at a frame rate of 30 Hz. The accuracy of the 3D measurement of Kinect v2 was validated in \cite{intro_4} where the measurement error is less than 0.07 cm in all three directions.

Several studies have been conducted in Kinect accuracy assessment, improvement and furthermore, design of  Kinect based cost-effective motion capture systems.  For instance, the hardware coefficient (such as, resolution, depth spacial precision, etc) and sensor accuracy (such as, average precision distribution and linearity, etc) were evaluated in  \cite{SO_1}, which clarifies the limitation of Kinect v2's depth measurement at the hardware level.
The accuracy of gait tracking through single Kinect was evaluated in \cite{intro_1}, which certifies the validation of a single Kinect to accurately measure lower extremity kinematics. Here, Kinect v2 was compared with a digital inclinometer  with the measurement error of less than $0.5\degree$ in sagittal and frontal plane rotation and less than $2\degree$ in transverse plane rotation. In addition, a hand kinematic analysis system is developed in \cite{intro_2}  which  measures the finger joint movement. By compared with Vicon system, the measurement absolute error is $2.4\degree$ which is better than clinically based alternative manual techniques in finger measurement. In \cite{intro_7}, an advanced hand gesture tracking interface was developed which can recognize 55 static and dynamic gestures with the accuracy of $92.4\%$. 

However, using a single Kinect sensor to develop an accurate postural tracking system is infeasible according to \cite{intro_3} and \cite{intro_6}. Compared with the accuracy of whole-body tracking result by Vicon, a single Kinect can not meet the accuracy requirement as a medical system. Since Microsoft released its software development kit (SDK) for Kinect v2, combining the measurements from multiple Kinects to improve the accuracy is possible. In \cite{SO_1}, a multiple Kinect v2 fusion approach was proposed based on linear least square principle, where the novelty was verified by comparing with the measurement from a single Kinect v2 sensor. Furthermore, in \cite{SO_2}, a geometric trilateration based motion capture system (with 3 Kinect v2 sensors) was developed, and a quantitative comparison with benchmark Delsys smart sensor system was conducted for gait tracking purpose. 


In this paper, a nonlinear least square trilateration method is proposed, based on which, a multiple Kinect motion capture system is developed.  Three key issues are addressed in this paper, i.e., calibration, occlusion compensation and accuracy improvement. Specifically, a calibration approach is designed to be easily set up and  be quickly configured. An occlusion compensation method is proposed to avoid the mistaken tracking due to the occlusion happened in the body movement. A nonlinear trilateration approach is proposed to optimize the tracking accuracy based on the redundant measurements of multiple Kinect v2 sensors located in different positions and oriented in different angles. And a synchronization method is proposed to synchronize clocks of clients and the server to organize measurements from sensors. The experiments are designed and conducted to evaluate the aforementioned four approaches, including examining the calibration approach by analyzing the factors affecting the Kinect's depth measurement, and evaluating the occlusion compensation approach by comparing the tracking result with and without occlusion compensation, validating the trilateration approach  by comparing the accuracy with the state-of-the-art: linear trilateration approach \cite{SO_1} and geometric trilateration approach \cite{SO_2}, and evaluating the efficiency by measuring the processing time of the frame and the clock error of the proposed system,  respectively. Compared with  \cite{SO_1}, the proposed nonlinear trilateration method has higher accuracy. Furthermore,  the accuracy of our proposed trilateration approach can be improved by adding more sensors while the geometric trilateration approach \cite{SO_2} is restricted as it only supports 3-Kinect setup. 

\section{Related Work}

Usually, optical motion capture system can be classified as marker-based and markerless-based.  Optical marker-based motion capture systems track markers placed on the human body where most of the commonly-used commercial systems  (e.g., Vicon and Optitrack) apply IR reflective markers although LED markers are also utilized in some specific marker-based systems \cite{rw_6}. On the other hand, Kinect can be considered as a markerless-based optical motion capture device. It applies human body structure principle \cite{rw_1} to recognize the human body against the background, and estimate the 3D coordinates of the body joints.

\subsection{Calibration}

To calibrate sensors, motion capture systems are required to measure the relative position of calibration points by transferring image points to the corresponding world coordinates \cite{rw_10}. Marker-based and markerless-based systems are supposed to focus different fields in calibration due to the fact that the markers can be accurately recognized from the background but the position of the markers can not be easily measured. Thus, marker-based systems always set up multiple markers \cite{rw_3,rw_4} for calibration by comparing the difference of relative position of markers from sensors, or compare the relative speed of moving markers \cite{rw_5}. The advantage of the marker-based systems' calibration is the stable accuracy that IR sensors are robust to the influence from the environment of the measurement area. However, the disadvantage is the calibration setup of the markers is always complex and time-consuming.

On the other hand, the markerless-based systems apply light coding or time-of-flight method to generate the depth map of its field-of-view, and use the RGB or depth data to recognize the calibration objects from the environment. Thus, the calibration object is required to be easily recognizable due to its special shape or optical property. At present, matrix barcode on the planar surface, i.e., quick response code (QR-Code), is the most wildly used technology on optical identification field. Open-source libraries for QR-code has been released by OpenCV which can be applied in various kinds of optical cameras, such as Optitrack \cite{rw_7}, Kinect \cite{intro_5},  Wii remotes \cite{rw_6}, etc. In addition, calibration objects can be designed with special shapes to be easily recognizable by depth sensors \cite{rw_8}\cite{rw_11}. Compared with marker-based calibration, the markerless-based calibration with QR-Code have a better time efficiency due to the fact that producing a QR-Code on a paper or surface is easy and efficient. In this paper, our proposed system applies the depth map to recognize the calibration object (details are illustrated in section \ref{sec:calibration}).

\subsection{Occlusion Compensation}

Occlusion is the primary factor affecting the accuracy of optical motion capture systems. Methods to detect occlusion and furthermore compensate its influence are quite different depending on different systems' structures. A single sensor can only obtain a side field-of-view, leading that positions of occluded joints can only be inferred. Thus, systems with single sensor are designed to apply likelihood algorithm to resolve the occlusion \cite{rw_9} which matches the measured depth data with human body models based on its movement restrictions. In this case, the system's accuracy is related to the likelihood estimation results. On the other hand, systems with multiple sensors can utilize the advantage of the distributed sensor layout: the occluded observations would be determined and furthermore ignored before the process of sensor data measurement. In this paper, the proposed occlusion approach applies the computation geometry principle which will be detailed explained in section \ref{sec:occlusion}.

\section{Overall System Architecture}

\begin{figure}[!t]
	\centering
	\includegraphics[width=9cm]{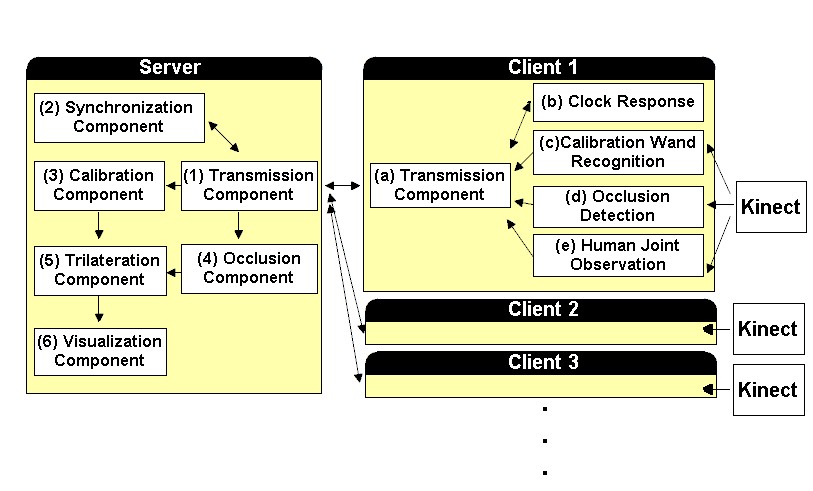}
	\par
	
	\protect\caption{\label{fig:architecture}Server-client network architecture of the proposed nonlinear trilateration motion capture system.}
\end{figure}

The proposed system's architecture is based on the server-client network structure (shown in Fig. \ref{fig:architecture}), which consists of a single server and several clients. The server controls all clients and compute the tracking results by applying the proposed nonlinear trilateration method. Furthermore, the server acts as the graphical user interface (GUI)  which receives user commands and displays the real-time tracking results. Each client connects with one Kinect v2 and continuously obtains the depth images which are sent to the server after the occlusion detection.  In Fig. \ref{fig:architecture}, the arrows in the figure indicate the transmission direction of data.

\subsection{Server Side}

The server has 6 components: 1) \textit{Transmission Component,} 2) \textit{Synchronization Component,} 3) \textit{Calibration Component, }4) \textit{Occlusion Compensation Component, }5)\textit{Trilateration Component,} and 6) \textit{Visualization Component,}. The details of these components are explained as follows:

1) \textit{Transmission Component:} The overall designed system is distributed leading that the transmission mechanism is critical. In addition, other components in the system work simultaneously where most of them need to exchange data through the network for fulfilling their functions. Here, the transmission component establishes and maintains the connection between the server and all clients that is the communication interface based on which, other components exchange messages. In order to achieve real-time, high accuracy, and robustness, it is important to guarantee the network transmission unimpeded. For that reason, we create multiple threads to execute different parallel missions, including categorizing, receiving, sending, and resending. The component is designed based on TCP/IP protocol and applies Winsocket APIs.





2) \textit{Synchronization Component:}  This component works once the connection established in the system by the Transmission Component. It dynamically calculates the difference of the clock between the server and clients, as well as the delay of the data transmission in the network; continuously sends messages nailed by time stamp to the client; and collects information of the network status and the client's clock. Specifically, this component is built based on the NTP protocol which was modified to fit the requirement of our system. Furthermore, this component sets up buffers for measurement data received from the clients. To obtain the results in real-time, the component restricts the interval of taking out data from the buffer, superfluous and independent data is removed.




3) \textit{Calibration Component:} This component registers the positions of sensors and computes the coordinate transformation for further trilateration (Component 5 in the server side). Kinect v2 is able to measure the depth map of its field-of-view, and transfer it into three-dimensional coordinates. The result is based on the local coordinate system from itself meaning that an integral coordinate system is needed to combine the measurements in local coordinates. In this respect, the calibration wand is applied to define the direction of axises and the origin point in the real world. 



4) \textit{Occlusion Compensation Component:} This component abandons the occluded tracking measurements according to the occlusion detection (Component d in the client side) from the numinous clients. Specifically, it dynamically bundles the tracking observations from Kinect sensors which are not fully/partially occluded by other body parts or objects for the further trilateration computation (Component 5 in the server side).


5) \textit{Trilateration Component:}  This component combines the results by applying nonlinear least square criterion to minimize the objective function which is defined as the difference between the computed joint trilateration results and the sensors' joint measurements.



6) \textit{Visualization Component:} This component is designed as the user interface on the server to show the trilateration results, i.e., real-time 2D projection of skeleton of the user's body on screen. Technically, this component is based on MFC (Microsoft Foudation Class) framework.  

\subsection{Client Side}

Each client has 5 components: a)  \textit{Transmission Component, }b)\textit{ Clock Response, }c) \textit{Calibration Wand  Recognition, }d) \textit{Occlusion Detection, }e) \textit{Human Joint Observation.}

The detailed functions of the mentioned components are illustrated as follows.  a) \textit{Transmission Component:} This component communicate with the server by sending the connection request, and then confirming the port and serial number through the response from the Transmission Component in the server side. b) \textit{Clock Response:} To synchronize the clock, the client receives the message sent by Synchronization Component (Component 2 in the server side), which responses this message with the current client's local time. c) {Calibration Wand  Recognition:} This component recognizes the calibration wand from the Kinect's depth map and localize it in the client's 3D coordinates for the Calibration Component (Component 3) in the server side.  d) \textit{Occlusion Detection:} The client pre-processes the measurement obtained from Kinect v2 to determine if occlusion  occurs, and sends the results to the Occlusion Component (Component 4) in the server side. e) \textit{Human Joint Observation:} The client applies Kinect APIs to recognize the human body joints from the measured depth map.

\section{Calibration Component}
\label{sec:calibration}
To achieve the designed purpose, digital image processing and trigonometry principles are applied in this component. The calibration wand is designed to be the origin of the server's coordinate system, and its trajectory defines the direction of axises. The basic idea of calibration is divided into three steps: first is to obtain the relative position of the calibration wand in the client's coordinate system (i.e., the local coordinate); second is to calculate the rotational angle and transform all position data from the client's coordinate system to the server's coordinate system.

\subsection{Calibration Wand Localization}

\begin{figure}[!t]
	\centering\includegraphics[width=7cm]{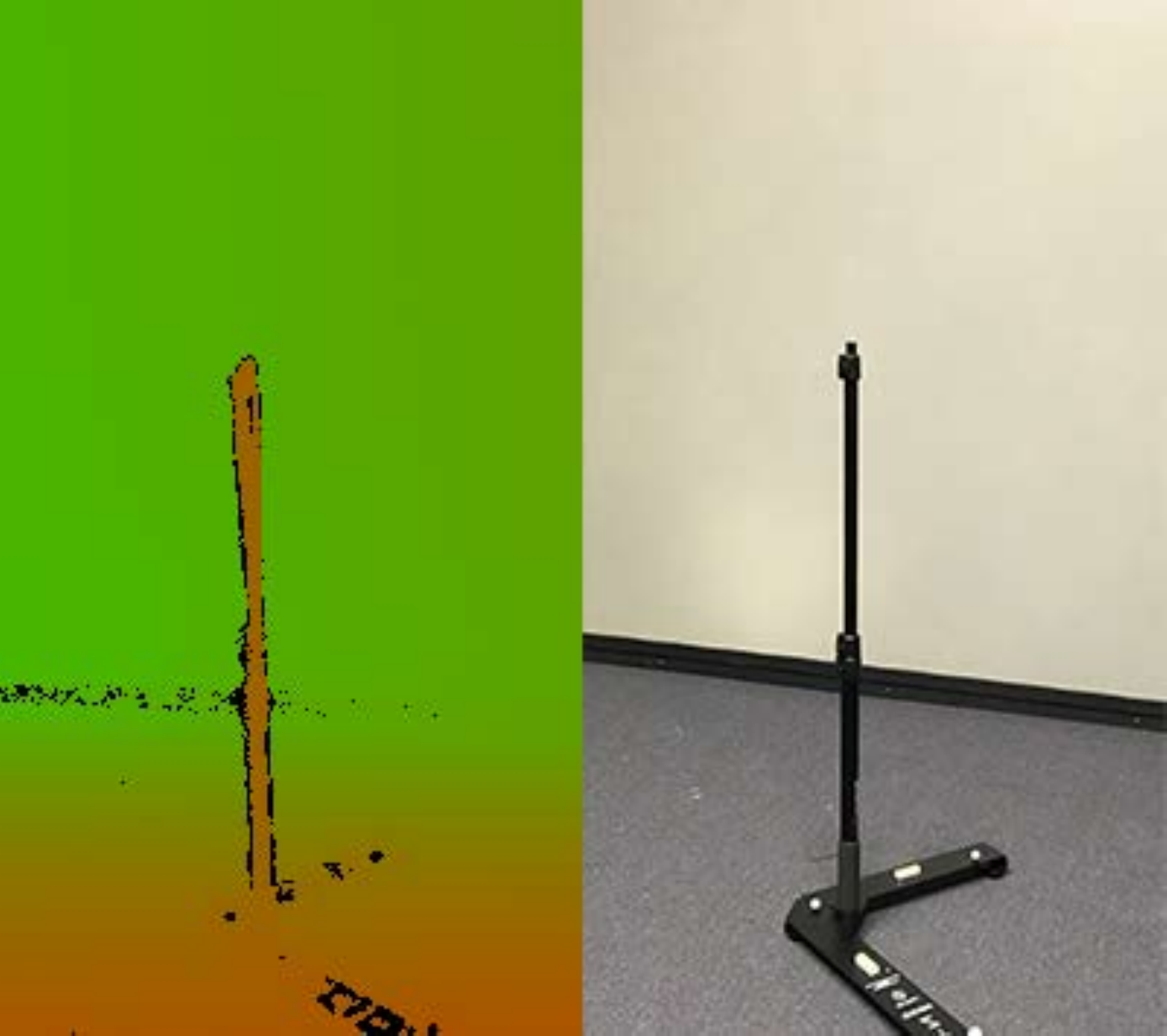}
	
	\protect\caption{\label{fig:depthimage1} Color image (left) and depth image (right) of the calibration wand in a mightily disturbed background. The black pixels in the depth image refers to the noise and shadow.}
\end{figure}

The main purpose of this step is to generate the area that covers the calibration wand from the whole 2D depth image for the further calculation.  The color image and the corresponding depth image of the calibration wand are shown in Fig.\ref{fig:depthimage1}. The bitmap of depth $D$ is defined to indicate the depth image obtained by Kinect where $D_{ij}$ refers the distance in centimeters between the Kinect and the corresponding point of object's surface. $i$ and $j$ indicate the horizontal and vertical coordinate of the depth image, respectively. As the designed detection range of the Kinect is from 0.5m to 6.5 m, object in the depth image exceeds such range results in inaccurate performance. Standardization of $D$ is conducted as the initialization:

\begin{equation}
\label{eq:calibration_bitmapstandardization}
D_{ij}=\left\{\begin{matrix}
50 & D_{ij}<50\\ 
650 & D_{ij}>650\\ 
D_{ij} & otherwise.
\end{matrix}\right.
\end{equation}

As shown in Fig.\ref{fig:depthimage1}, pixels constitute the image of the wand are orange. To obtain the data from the image, iterated binarization procedure is conducted to differentiate the image pixels of the calibration wand from the background. The labelmap $T$ is defined to indicate the result of the binarization where $T_{ij}=1$ indicates that $D_{ij}$ is labeled as the calibration wand's image pixel. In each iteration, $T_{ij}^{k}$ (the element in the labelmap with coordinate $(i,j)$ in the $k$-th iteration) is labeled as 1 when $D_{ij}$ is no more than the threshold $R^{k}$ which is computed as

\begin{equation}
\label{eq:calibration_bitmapbinarization1}
R^{k}=  Ave\left(D_{ij}|T^{k-1}_{ij}=1 \right)\cdot C_{offset}\\ 
\end{equation}
where all elements in the initial labelmap, i.e, $T^{1}$ are labeled as 1. $C_{offset}$ indicates the parameter to avoid the occurred offset, and  $Ave\left (\bullet \right )$ refers to the average of the set $\left (\bullet \right )$. In this paper, the coefficient  $ C_{offset}$ is chosen as 0.5025.

The iteration ends if the condition (Equation   \ref{eq:calibration_bitmapbinarization3}) is satisfied. It means that the maximum difference of the depth of labeled pixels is less than the $C_{offset}$. Thus. only the wand's image is labeled.

 \begin{equation}
 \label{eq:calibration_bitmapbinarization3}
 \begin{matrix}
 \mathrm{Min}(D_{ij}|T^{k}_{ij}=1 )\cdot C_{offset}>=\mathrm{Max}(D_{ij}|T^{k}_{ij}=1)
 \end{matrix}
 \end{equation}

According to $T$, the boundary of the calibration wand in the depth image is determined by the difference of radius. Therefore the upper left point and the bottom right point of the image are defined as $P_{ul}$ and $P_{br}$, respectively. The coordinate of the center point of the calibration wand's image can be calculated through these two points which are defined as $P_{m}$.
According to the technical specification \cite{inrtro_kinect_1}, the range of viewing field of Kinect v2 is $70^{\circ}$ and $60^{\circ}$ in horizontal and vertical directions, respectively. As the size of depth image is $512\times424$ pixels, the central point of the image $P_{c}$ is $\left(256,212 \right)$. The 3D coordinate of a point in the real world (denoted as $P' (x_{P'}, y_{P'},z_{P'})$), corresponding with the point in the depth image (denoted as $P(x_{P}, y_{P},z_{P})$), are calculated as


\begin{equation}
\label{eq:calibration_2dlocation}
\left\{\begin{array}{cl}
x_{P'}=& D_{P}\cdot tan(70^{\circ}\cdot\frac{x_{P}-256}{512})\\ 
y_{P'}=& D_{P}\cdot tan(60^{\circ}\cdot\frac{y_{P}-212}{424})\\ 
z_{P'}=& D_{P}
\end{array}\right.
\end{equation}


However, the origin of the server's coordinate system is the center of the calibration wand instead of the center of the calibration wand's surface. Therefore, once the coordinate of the point $P'_{m}(x_{P'_{m}}, y_{P'_{m}},z_{P'_{m}})$ corresponding to  $P_{m}$ is calculated, the radius of calibration wand (i.e., $x_{P_{br}}-x_{P_{ul}}$) need to be taken into consideration as well. The coordinate of the center of calibration wand in the coordinate system of client $P'_{o}(x_{P'_{o}}, y_{P'_{o}},z_{P'_{o}})$ are calculated as

\begin{equation}
\label{eq:calibration_3dlocation}
\left\{\begin{array}{cl}
x_{P'_{o}}=& x_{P^{`}_{m}}\\ 
y_{P'_{o}}=& y_{P^{`}_{m}}\\ 
z_{P'_{o}}=& z_{P^{`}_{m}}+(x_{P_{br}}-x_{P_{ul}})
\end{array}\right.
\end{equation}

 To enhance the robustness of this calibration wand recognition, the client sensor obtains measurement several times in each calibration frame and compares the difference of the measurement results. The frame with a higher deviation of measurement is ignored.

\subsection {Transformation from Client's Coordinates to Server's Coordinates}

\begin{figure}[!t]
	\centering\includegraphics[width=8cm]{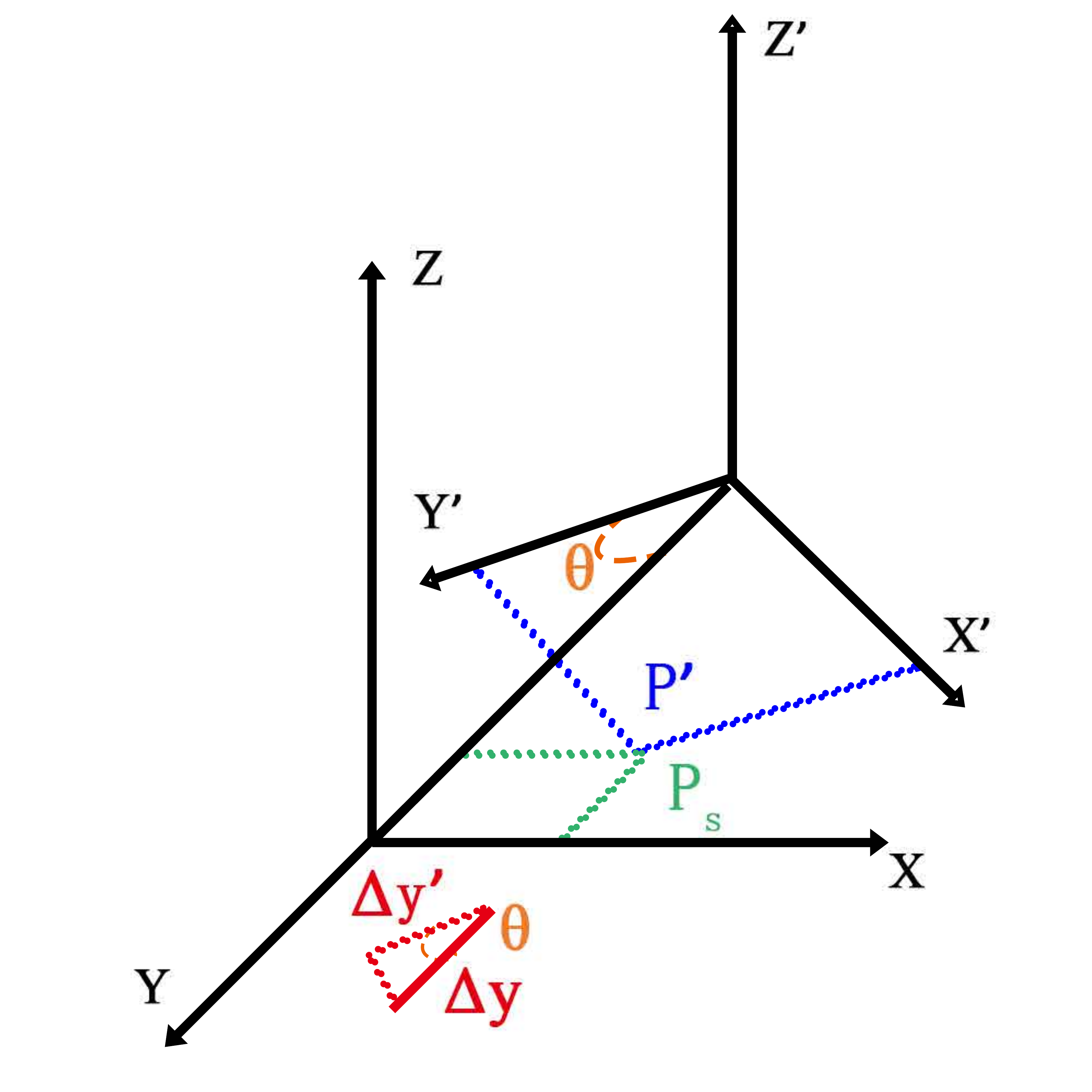}
	
	\protect\caption{\label{fig:coordinatetransformation} 
	Defining $\theta$, $P_{s}$,  $P'$, $\Delta y$ and $\Delta y'$ in the transformation from the client's coordinate to the server's coordinate.}
\end{figure}

In our proposed method, all  Kinects are located to be perpendicular to the ground surface, meaning that z-axises of all coordinates (including server's and clients') are parallel. Therefore, there is no rotation transformation along z-axis. The rotational angles are calculated by the differences between the coordinate movements (in x-axis and y-axis directions) within sever's and client's coordinate systems.

The client obtains the position of calibration wand twice which defines a trajectory of the calibration wand. Here, we refer this trajectory as the y-axis in the server's coordinates. The rotational angle $\theta$ of the  y-axis can be calculated as

\begin{equation}
\label{eq:rotataionangle}
\theta=arccos \left(\frac{\Delta y'}{\Delta y}\right)
\end{equation}
where $\Delta$ refers to change operator. $ y'$ and $y$ indicate the y coordinate of the calibration wand in the client's coordinates and server's coordinates, respectively.

With relative displacement of origins and the rotation angle of axises, a point in the client's coordinate ($P'(x_{P'}, y_{P'},z_{P'})$) can be transformed to its corresponding point in the server's coordinate ($P^{s}(x_{P^{s}}, y_{P^{s}},z_{P^{s}})$) as

\begin{equation}
\label{eq:coordinatetransformation}
\left\{\begin{array}{cl}
x_{P^{s}}=& x_{P'}\cdot cos(\theta)-y_{P'}\cdot sin(\theta)\\ 
y_{P^{s}}=& x_{P'}\cdot sin(\theta)+y_{P'}\cdot cos(\theta)\\ 
z_{P^{s}}=& z_{P'}
\end{array}\right.
\end{equation}

\section{Occlusion Compensation Component}
\label{sec:occlusion}



Kinect v2 is designed to obtain inferred results when it is not able to make the direct measurement, leading that the accuracy of the measurement is not accurate enough compared with directly tracked results.  The inferred results are first marked by the Kinect's SDK. In Addition, the client analyzes the position of limbs to determine if the results are affected by occlusion.  

As the image of the human body is abstracted to combined segments in 3D coordinates, occlusion of limbs in this paper is defined as the crossing of segments on the projection in Kinect's view. Vector product from computational geometry principle is applied to determine the crossing of segment $V_{1}$ and $V_{2}$. Two segments are crossed when 6 conditions in Equation \ref{eq:segmentcross1} are met.
 
\begin{equation}
\label{eq:segmentcross1}
\left\{\begin{array}{lll}
\multicolumn{2}{l}{\mathrm{Min}(x_{V_{1},1}, x_{V_{1},2}) - \mathrm{Max}(x_{V_{2},1}, x_{V_{2},2})} &<=0  \\ 
\multicolumn{2}{l}{\mathrm{Min}(x_{V_{2},1}, x_{V_{2},2}) - \mathrm{Max}(x_{V_{1},1}, x_{V_{1},2})}&<= 0 \\
\multicolumn{2}{l}{\mathrm{Min}(y_{V_{1},1}, y_{V_{1},2}) -\mathrm{Max}(y_{V_{2},1}, y_{V_{2},2})}& <=0  \\
\multicolumn{2}{l}{\mathrm{Min}(y_{V_{2},1}, y_{V_{2},2}) -\mathrm{Max}(y_{V_{1},1}, y_{V_{1},2})}& <= 0\\
VectorProduct&( x_{V_{1},1} - x_{V_{2},1}, y_{V_{1},1} - y_{V_{2},1}, \\ &\ x_{V_{2},2} - x_{V_{2},1}, y_{V_{2},2} - y_{V_{2},1})\cdot\\
VectorProduct&( x_{V_{1},2} - x_{V_{2},1}, y_{V_{1},2} - y_{V_{2},1}, \\ &\ x_{V_{2},2} - x_{V_{2},1}, y_{V_{2},2} - y_{V_{2},1}) &<  0 \\
VectorProduct&( x_{V_{2},1} - x_{V_{1},1}, y_{V_{2},1} - y_{V_{1},1}, \\ &\ x_{V_{1},2} - x_{V_{1},1}, y_{V_{1},2} - y_{V_{1},1})\cdot\\
VectorProduct&( x_{V_{2},2} - x_{V_{1},1}, y_{V_{2},2} - y_{V_{1},1}, \\ &\ x_{V_{1},2} - x_{V_{1},1}, y_{V_{1},2} - y_{V_{1},1}) &<  0\\

\end{array}\right.
\end{equation}
where $x_{V_{i},j}$ indicates the $x$ coordinate of the $j$-th point's coordinate in the segment $V_{i}$, similarly to $y_{V_{i},j}$ and $z_{V_{i},j}$, respectively. $VectorProduct(x_{1},x_{2}, y_{1},y_{2})$ refers to the vector product operator which is $(x_{1}\cdot y_{2}-x_{2}\cdot y_{1})$.

It is noted that two crossed segments are intersected at a single point unless they are overlapped. Furthermore, the sequence of crossed segments is determined by the spacial distance to the Kinect sensor (i.e., the coordinate $z$) of the intersection point. With the result of the vector product of two segments, the coordinate of intersection point is calculated, and the sequence of them can be further obtained. Therefore, the occluded limbs are determined by calculating the intersection points of all pairs of segments in the image of the human body, and comparing the $z$ coordinate of the intersection point. The occluded data is marked by the client before sending to server. Here, only data without occlusion is transferred to the next procedure if at least one set of trusted data exists; otherwise, the data with minimum number of intersection points is selected.

\section{Nonlinear Trilateration Component}
\label{sec:trilateration}

This component is to combine the measurements  from each sensor to obtain a more accurate trilateration result \cite{trilateration_1}. The coordinate transformation (Section \ref{sec:calibration}) is conducted through the result from calibration component before the measurement is applied to the nonlinear trilateration calculation. Once it is done, this trilateration component starts the process that server continuously receives measurements from clients which contain the coordinates of the human body's joints. 

Specifically, the nonlinear least square method calculates an accurate position estimation with approximately accurate input measurements. First, the estimation error of the current trilaterated joint coordinate  $P(x_{P},y_{P},z_{P})$ with respect to the $i$-th sensor $f_{i}(P)$  is define as

\begin{equation}
\label{eq:objectivefunction2}
f_{i}(P)=\sqrt{   \left(x_{P}-x_{i}\right)^2+\left(y_{P}-y_{i}\right)^2+\left(z_{P}-z_{i}\right)^2}-r_{i}
\end{equation}
where $(x_{i},y_{i},z_{i})$ denotes  the coordinate of the $i$-th sensor, and $r_{i}$ denotes to the distance between the current estimated joint position and the $i$-th sensor's position, which would not be measured or estimated as its purpose is for deriving the partial differentiations in Eq.\ref{eq:partialderivative}. Here, an objective function $F$ is defined as the sum of the square of the estimation errors.  To obtain an optimized joint position by nonlinear trilateration is equivalently to solve the following minimization problem regarding with  $F$:

\begin{equation}
\label{eq:objectivefunction}
\min F(P)=\sum_{i=1}^{n}f_{i}(P)^{2}
\end{equation}

Considering the balance between the complexity and accuracy, Newton iteration is chosen \cite{trilateration_1} to solve the optimization problem in Equation \ref{eq:objectivefunction}.
More specifically, the partial derivative of the objective function $F$ with respect to $x$, $y$, $z$ yields

\begin{equation}
\label{eq:partialderivative}
\left\{\begin{array}{ll}
\frac{\partial F}{\partial x}=& 2\sum_{i=1}^{n}f_{i}(\frac{\partial f_{i}}{\partial x})\\

\frac{\partial F}{\partial y}=& 2\sum_{i=1}^{n}f_{i}(\frac{\partial f_{i}}{\partial y})\\

\frac{\partial F}{\partial z}=& 2\sum_{i=1}^{n}f_{i}(\frac{\partial f_{i}}{\partial z})
\end{array}\right.
\end{equation}

Referred to Newton iteration, vectors $\vec{g}$ and $\vec{R}$ are defined

\begin{equation}
\label{eq:newton1}
\begin{matrix}
\underbrace{
\begin{pmatrix}
\frac{\partial F}{\partial x}\\
\frac{\partial F}{\partial y}\\
\frac{\partial F}{\partial z}
\end{pmatrix}}_{\vec{g}}
=2
\underbrace{
\begin{pmatrix}
\frac{\partial f_{1}}{\partial x} & \frac{\partial f_{1}}{\partial y} & \frac{\partial f_{1}}{\partial z}\\ 
\frac{\partial f_{2}}{\partial x} & \frac{\partial f_{2}}{\partial y} & \frac{\partial f_{2}}{\partial z}\\ 
\vdots & \vdots  & \vdots \\ 
\frac{\partial f_{n}}{\partial x} & \frac{\partial f_{n}}{\partial y} & \frac{\partial f_{n}}{\partial z}
\end{pmatrix}^{\mathrm{T}}}_{J}
\underbrace{
\begin{pmatrix}
f_{1}\\ 
f_{2}\\ 
\vdots\\
f_{n}
\end{pmatrix},
}_{\vec{f}}
\end{matrix}
\end{equation}
and
\begin{equation}
\label{eq:vectorrdefinition}
\vec{R}=\begin{pmatrix}
x_{P}\\
y_{P}\\
z_{P}
\end{pmatrix}
\end{equation}

Assuming the intermediate solution of $\vec{R}$ in the $k$-th iteration is noted as $\vec{R}_{k}$.  According to Newton iteration, the the solution in the next iteration $\vec{R}_{k+1}$ can be computed as

\begin{equation}
\label{eq:newtoniteration1}
\vec{R}_{k+1}=\vec{R}_{k}-\left( J^{\mathrm T}_{k} J_{k}  \right)^{-1}J^{\mathrm T}_{k}\vec{f}_{k}
\end{equation}

The initial solution $\vec{R}_{1}$ can be obtained by the linear least square method \cite{SO_1}. Regarding the iteration rule in Equation \ref{eq:newtoniteration1}, on the right side, we provide the calculation of $J^TJ$ and $J^{\mathrm T}\vec{f}$ as follows

\begin{equation}
\label{eq:newtoniteration2}
J^{\mathrm{T}}J=
\begin{pmatrix}
A_{1} & A_{2} & A_{3}
\end{pmatrix}
\end{equation}
where
\begin{equation}
\begin{matrix}
A_{1}=\begin{pmatrix}
\sum_{i=1}^{n}\frac{\left(x_{P}-x_{i} \right )^{2}}{\left(f_{i}+r_{i} \right )^{2}}\\
\sum_{i=1}^{n}\frac{\left(x_{P}-x_{i} \right )\left(y_{P}-y_{i} \right)}{\left(f_{i}+r_{i} \right )^{2}}\\
\sum_{i=1}^{n}\frac{\left(x_{P}-x_{i} \right )\left(z_{P}-z_{i} \right)}{\left(f_{i}+r_{i} \right )^{2}}

\end{pmatrix}\\
A_{2}=\begin{pmatrix}
\sum_{i=1}^{n}\frac{\left(x_{P}-x_{i} \right )\left(y_{P}-y_{i} \right)}{\left(f_{i}+r_{i} \right )^{2}}\\
\sum_{i=1}^{n}\frac{\left(y_{P}-y_{i} \right )^{2}}{\left(f_{i}+r_{i} \right )^{2}} \\
\sum_{i=1}^{n}\frac{\left(y_{P}-y_{i} \right )\left(z_{P}-z_{i} \right)}{\left(f_{i}+r_{i} \right )^{2}}
\end{pmatrix}\\
A_{3}=\begin{pmatrix}
\sum_{i=1}^{n}\frac{\left(x_{P}-x_{i} \right )\left(z_{P}-z_{i} \right)}{\left(f_{i}+r_{i} \right )^{2}}\\
\sum_{i=1}^{n}\frac{\left(y_{P}-y_{i} \right )\left(z_{P}-z_{i} \right)}{\left(f_{i}+r_{i} \right )^{2}}\\
\sum_{i=1}^{n}\frac{\left(z_{P}-z_{i} \right )^{2}}{\left(f_{i}+r_{i} \right )^{2}}
\end{pmatrix}
\end{matrix}
\end{equation}
and
\begin{equation}
\label{eq:newtoniteration3}
J^{\mathrm T}\vec{f}=
\begin{pmatrix}
\sum_{i=1}^{n}\frac{\left(x_{P}-x_{i} \right )f_{i}}{\left(f_{i}+r_{i} \right )^{2}}\\
\sum_{i=1}^{n}\frac{\left(y_{P}-y_{i} \right )f_{i}}{\left(f_{i}+r_{i} \right )^{2}}\\
\sum_{i=1}^{n}\frac{\left(z_{P}-z_{i} \right )f_{i}}{\left(f_{i}+r_{i} \right )^{2}}
\end{pmatrix}
\end{equation}

Since $J^{\mathrm T}J$ is a matrix with a size of 3 by 3, adjugate matrix method is applied to calculate its inverse matrix to improve the efficiency as

\begin{equation}
\label{eq:matrixinverse1}
\left( J^{\mathrm T}J \right)^{-1} = \frac{\left( J^{\mathrm T}J \right)'}{\left |  J^{\mathrm T}J \right |}
\end{equation}
where $\left( J^{\mathrm T}J \right)'$ indicates the adjugate matrix which is the transpose of the cofactor matrix of $J^{\mathrm T}J$ and $\left |  J^{\mathrm T}J \right |$ denotes the determinant of  $J^{\mathrm T}J$.

Finally, the iteration is designed to stop when the two adjacent iterative result of the objective function $F(P)$ is small enough, i.e.,

\begin{equation}
\label{eq:iterationend}
F^{k}(P)-F^{k-1}(P)<C_{threshold}
\end{equation}
where $F^{k}(P)$ and $F^{k-1}(P)$ denote the result of the objective function in the $k$-th and $(k-1)$-th iteration, respectively. $C_{threshold}$ denotes the threshold of the adjacent iterative result.

In this paper, the experiment to determine the  $C_{threshold}$ is conducted where the trilateration input (i.e., 4 Kinect sensor measurements) is simulated with different errors and thresholds to test the time efficiency and accuracy. Each set of experiment takes 5000 sets of data that the standard deviation of the result is less than $0.001\%$. The result of experiments is shown in Table \ref{table:iterationexperiment}.
\begin{table*}
	
	\centering
	\protect\caption{\label{table:iterationexperiment} Compromise of Efficiency and Accuracy with Different Measurement Error Ranges and $C_{threshold}$}
	\begin{tabular}{|>{\centering}p{1.5cm}|>{\centering}p{1.5cm}|>{\centering}p{1.3cm}|>{\centering}p{1.5cm}|>{\centering}p{1.3cm}|>{\centering}p{1.5cm}|>{\centering}p{1.3cm}|}
		\hline
		\multirow{3}{*}{$C_{threshold}$}&
		\multicolumn{6}{c|}{Measurement Error Range}
		\tabularnewline\cline{2-7}
		&
		\multicolumn{2}{c}{ (15, 30] cm} &
		\multicolumn{2}{|c|}{[5, 15] cm} &
		\multicolumn{2}{c|}{[0, 5) cm} 
		\tabularnewline\cline{2-7}

		&
		Propagation Error (cm)&
		Consuming Time ($\mu$s)&
		Propagation	Error (cm)&
		Consuming Time ($\mu$s)&
		Propagation	Error (cm)&
		Consuming Time ($\mu$s)
		\tabularnewline\hline
		
		1&
		21.4&
		$<0.1$&
		17.2&
		$<0.1$&
		9.3&
		$<0.1$
		\tabularnewline\hline
		
		$1\e{-2}$&
		15.7&
		$0.6$&
		12.0&
		$0.4$&
		7.5&
		$0.3$
		\tabularnewline\hline            
		
		$1\e{-4}$&
		14.2&
		$11$&
		7.3&
		$10$&
		3.9&
		$7$
		\tabularnewline\hline                        
		
		$1\e{-6}$&
		13.7&
		$1040$&
		6.6&
		$783$&
		3.6&
		$651$
		\tabularnewline\hline    
		
	\end{tabular}
\end{table*}
 According to the result, with $C_{threshold}$ decreases, the accuracy improves and  the consuming time increases. When $C_{threshold}$ is set to $1\e{-6}$, compared with the case when  $C_{threshold}$ is $1\e{-4}$, the accuracy slightly improves, whereas the consuming time apparently  increases. For instance, the consuming time is more than 1 millisecond when the mean measurement error is bigger than $15$ cm. Thus, in this paper, $C_{threshold}$ is chosen as $1\e{-4}$.

-\section{Synchronization Component}
As the proposed system is based on LAN (Local Area Network), the time delays of mutual transmission are considered equal. 

This server periodically sends synchronization message to clients (in the server time $T_{1}$). When a client receives a synchronization message (in the client time $T_{2}$), it replies immediately (in the client time $T_{3}$). After that, the server receives the response (in server time $T_{4}$) of the synchronization message. The clock error and time delay between the server and client can be computed as 

\begin{equation}
\label{eq:synceq1}
\left\{\begin{matrix}
T_{2}-T_{1}=D_{i}+E_{i}\\ 
T_{4}-T_{3}=D_{i}+E_{i}\\ 
\end{matrix}\right.
\end{equation}
where $E_{i}$ denotes the clock error between the $i$-th client with the server, and $D_{i}$ indicates the time delay in transferring the data from the $i$-th client to the server. Derived from Equation \ref{eq:synceq1}, $D_i$ and $E_i$ can be computed as

\begin{equation}
\label{eq:synceq2}
\left\{\begin{array}{ll}
D_{i}=\frac{\left(T_{4}-T_{1}\right)-\left(T_{3}-T_{2}\right)}{2}\\ 
E_{i}=\frac{\left(-T_{2}+T_{1}\right)+\left(T_{4}-T_{3}\right)}{2}
\end{array}\right.
\end{equation}

Thus, the server can backtrack the sending time $T_{s,i}$ of a message (from the $i$-th client) based on its receiving time $T_{r,i}$,  $D_{i}$, and $E_{i}$ as

\begin{equation}
\label{eq:senttime}
T_{s,i}=T_{r,i}-D_{i}-E_{i}
\end{equation}

All received data is stored in buffers which are FIFO (First in first out) sequences categorized with the serial number of departure client. TCP/IP protocol would protect the sequence in time of data streams in the local network. However, it occasionally happens that some of the data lost during the transmission. In this case, the sending time is verified when the trilateration component extracts data from the buffer. In each frame of the server, a time range is determined by adding the extra offset time (i,e, $\frac{1000}{30}$ ms in fps of 30) into the current server time where the data exceeding the mentioned time range are excluded. 

\section {Results}

\subsection{Experiment Setup}

\begin{figure*}
    \centering
	\subfloat[][a]{\centering\includegraphics[width=9cm]{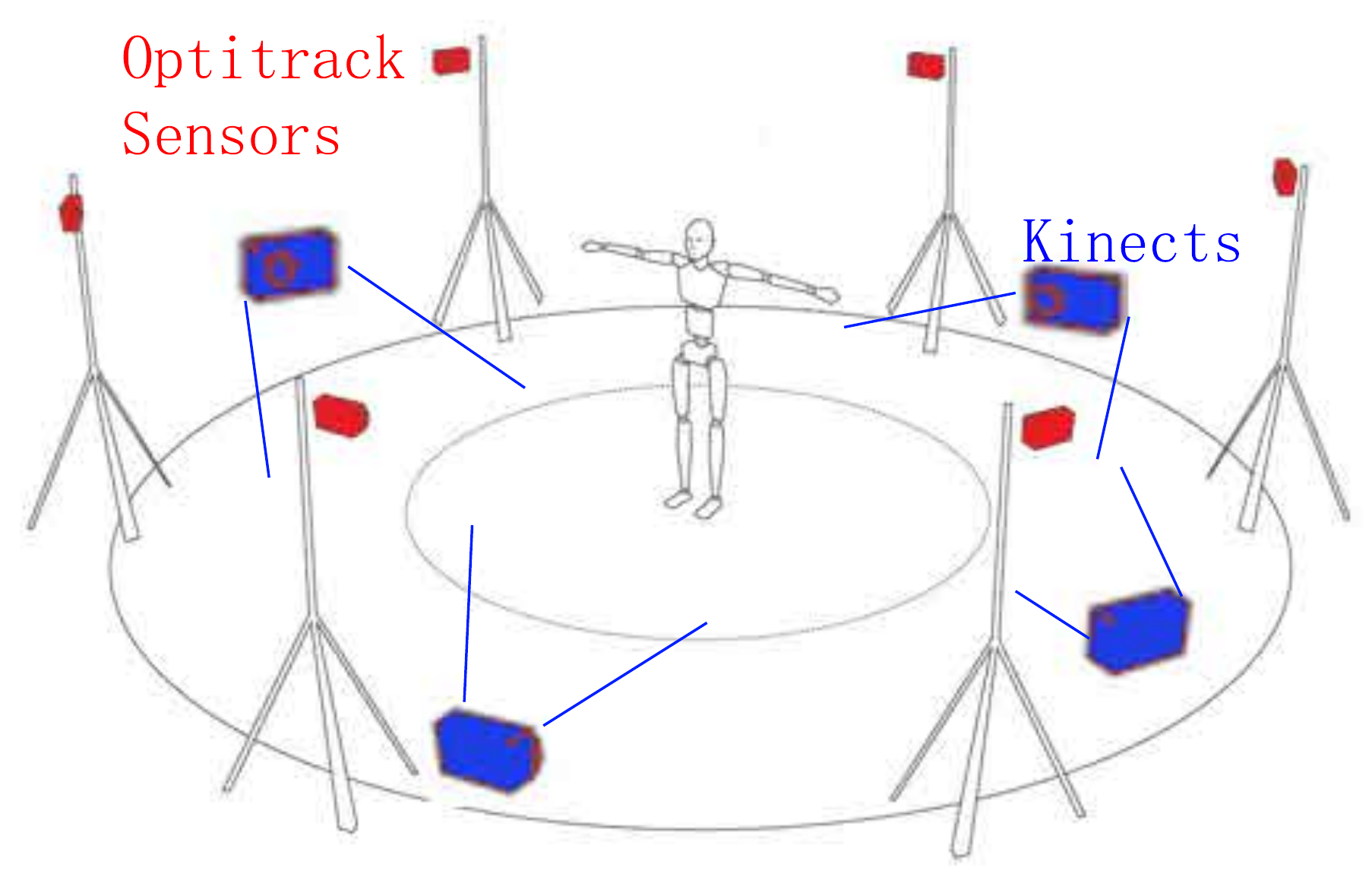}}	
	\subfloat[][b]{\centering\includegraphics[width=9cm]{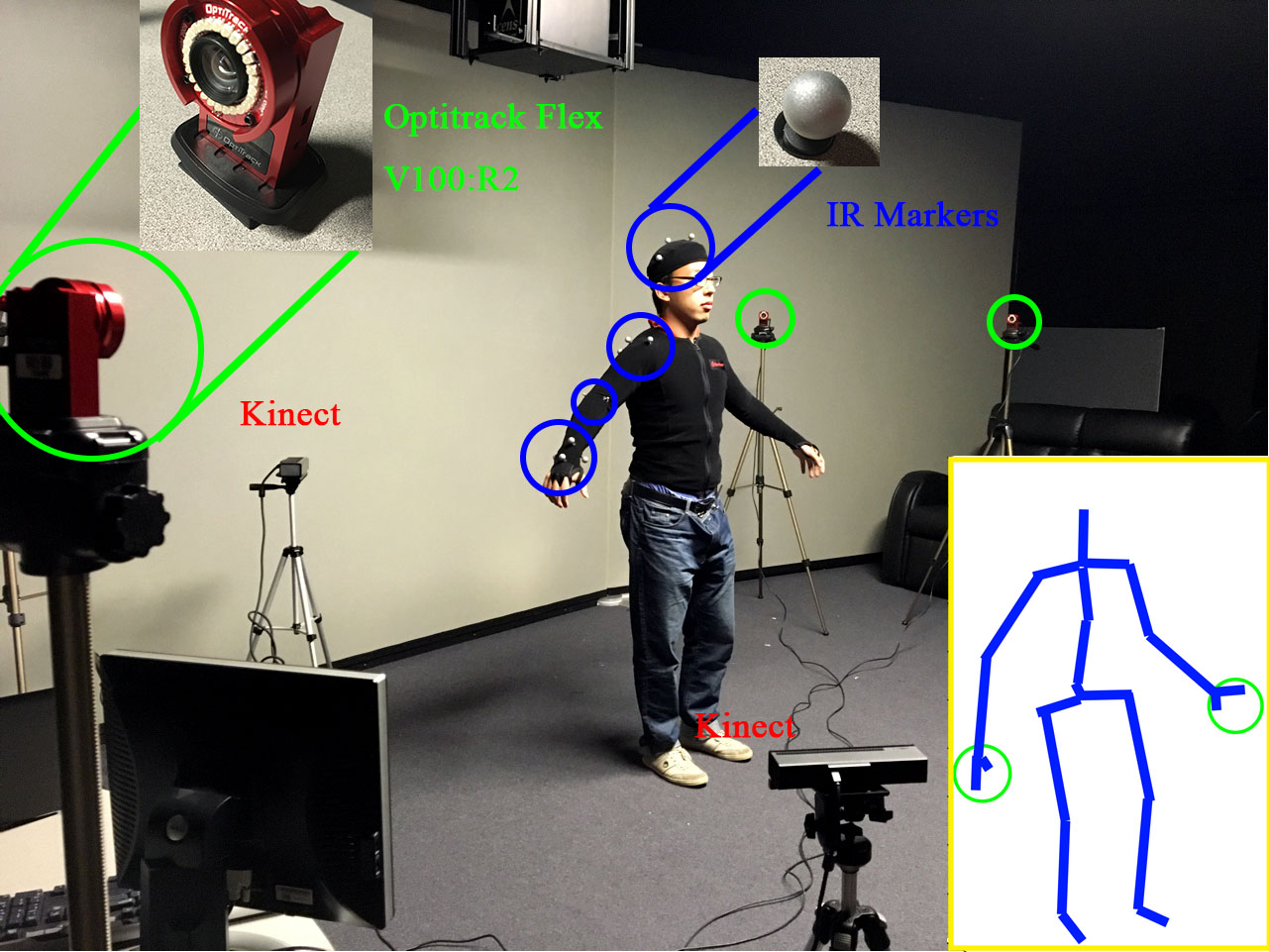}}
    \label{fig:experiment_photo}
    \caption{Experiment set-up. (a) The layout of Optitrack and Kinect v2 sensors in the experiment. (b) Snapshot of the experiment where the green and blue circles indicate the Optitrack sensors and IR markers that are placed on the tracked human joints. The visualization component's output is shown at the right bottom corner where the blue lines indicate the limbs and green circles indicate the hand tracking.}
\end{figure*}

Optitrack is a marker-based optical motion capture system with multiple IR camera sensors. It is chosen to be the ground-truth in the experiment for its high accuracy (mean absolute error < 1 mm ) \cite{optitrackr2}. Specifically, Optitrack Flex V100:R2 is the camera type, and Arena v1.7 is the corresponding software's version. Six Optitrack R2 cameras are placed around the measurement area with a radius of 2 m (the best measurement range for both Optitrack R2 and Kinect v2). All cameras are connected by a USB-Hub and controlled by a central computer. 4 Kinects (i.e., clients) are placed around the measurement area and a laptop computer (Processor: Intel(R) Core(TM) i7-4700MQ CPU @ 2.40GHZ, RAM: 8.00GB) is chosen as the server. In Addition, clocks from the server of the proposed system and the Optitrack are synchronized before each set of experiments to guarantee the clock error is less than 5 ms. The experiment setup and corresponding real-time tracking visualization is shown in Fig.  \ref{fig:experiment_photo} where (a) shows the layout of Optitrack and Kinect v2 sensors. (b) is the snapshot of the experiment scene as well as the real-time tracking visualization result. Totally 16 participants with height of $170\pm17$ cm and weight of $76.5\pm23.5$ kg  participated in the following experiments. As the requirement of Optitrack, all participants wear the pure black body suit with 4 IR markers placed in each tracked joints during the measurement.
 
\subsection{Calibration Evaluation}

\begin{figure}[!t]
	\centering
	\includegraphics[width=9cm]{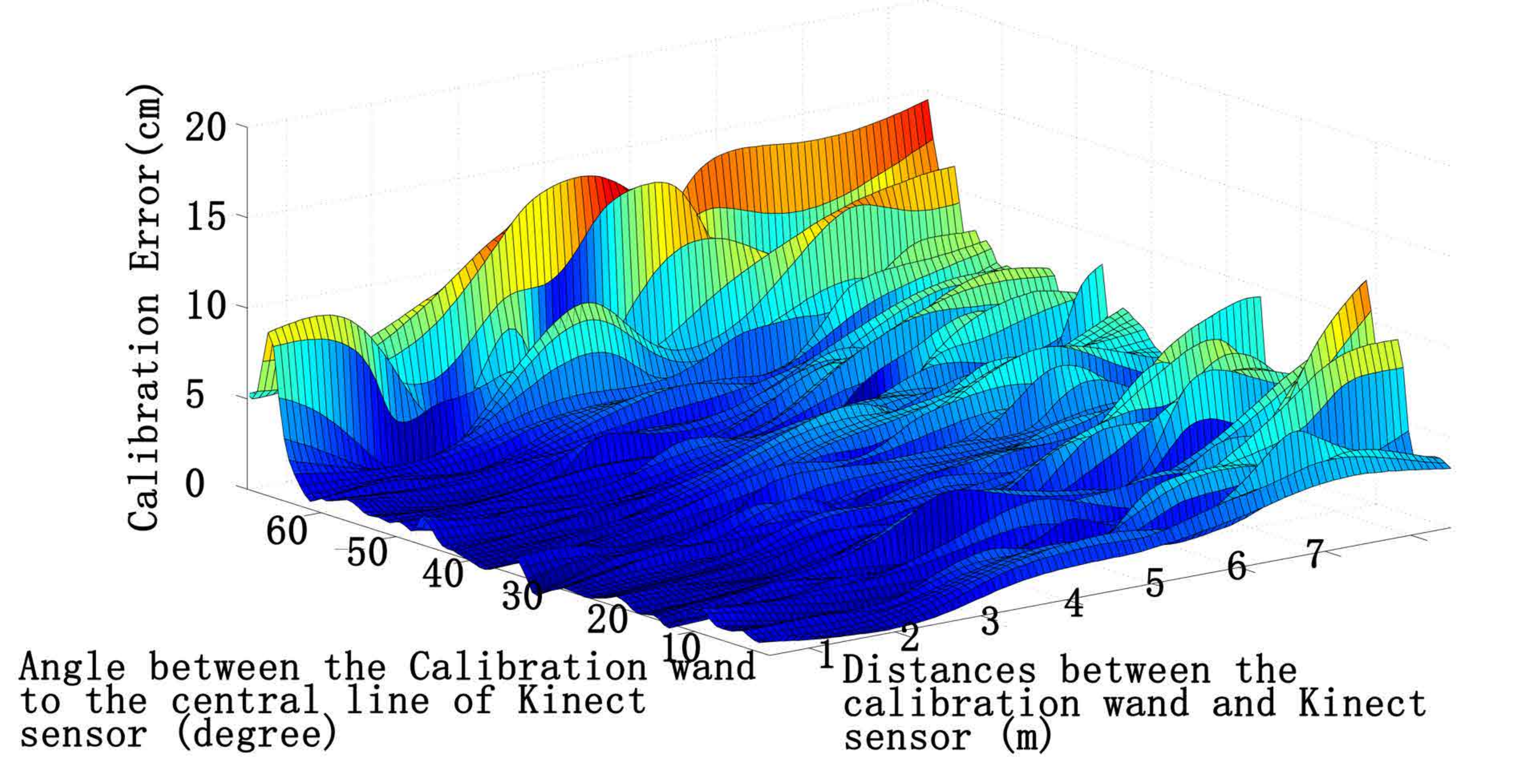}
	\par
	
	\protect\caption{\label{fig:calibration_reuslt}Calibration error distribution along the distance (between  the calibration wand and the Kincect sensor) and the angle (between the calibration wand and the central line of the Kinect sensor).}
\end{figure}

The accuracy of the calibration component is tested by comparing the measurement between the proposed system and the laser meter. The calibration results are based on different distances between the calibration wand and a sensor as well as the angle between the calibration wand to the central line of Kinect sensor. Here, 100 data samples are collected. The relation between the calibration error and the aforementioned distance and angle is shown in Fig. \ref{fig:calibration_reuslt}.

The specification of Kinect v2 provides its best measuring range which is 0.5 m - 3 m, and the obtained experiment results coincide with it. As shown in Fig. \ref{fig:calibration_reuslt},  the calibration error apparently increases if the calibration wand is very close to the sensor (i.e. less than 0.8 m) . The sensor can hardly conduct the calibration if the distance is less than 0.5 m due to the fact that there is only $20\%$ of calibration results are valid. On the other hand, if the distance between the wand and sensor is more than 3 m, the calibration error apparently increases because of the resolution limitation of the Kinect v2 sensor. To conclude, the distance between the calibration wand and sensors should be kept between 0.8 m and 3 m. 

\begin{figure}[!t]
	\centering
	\includegraphics[width=9cm]{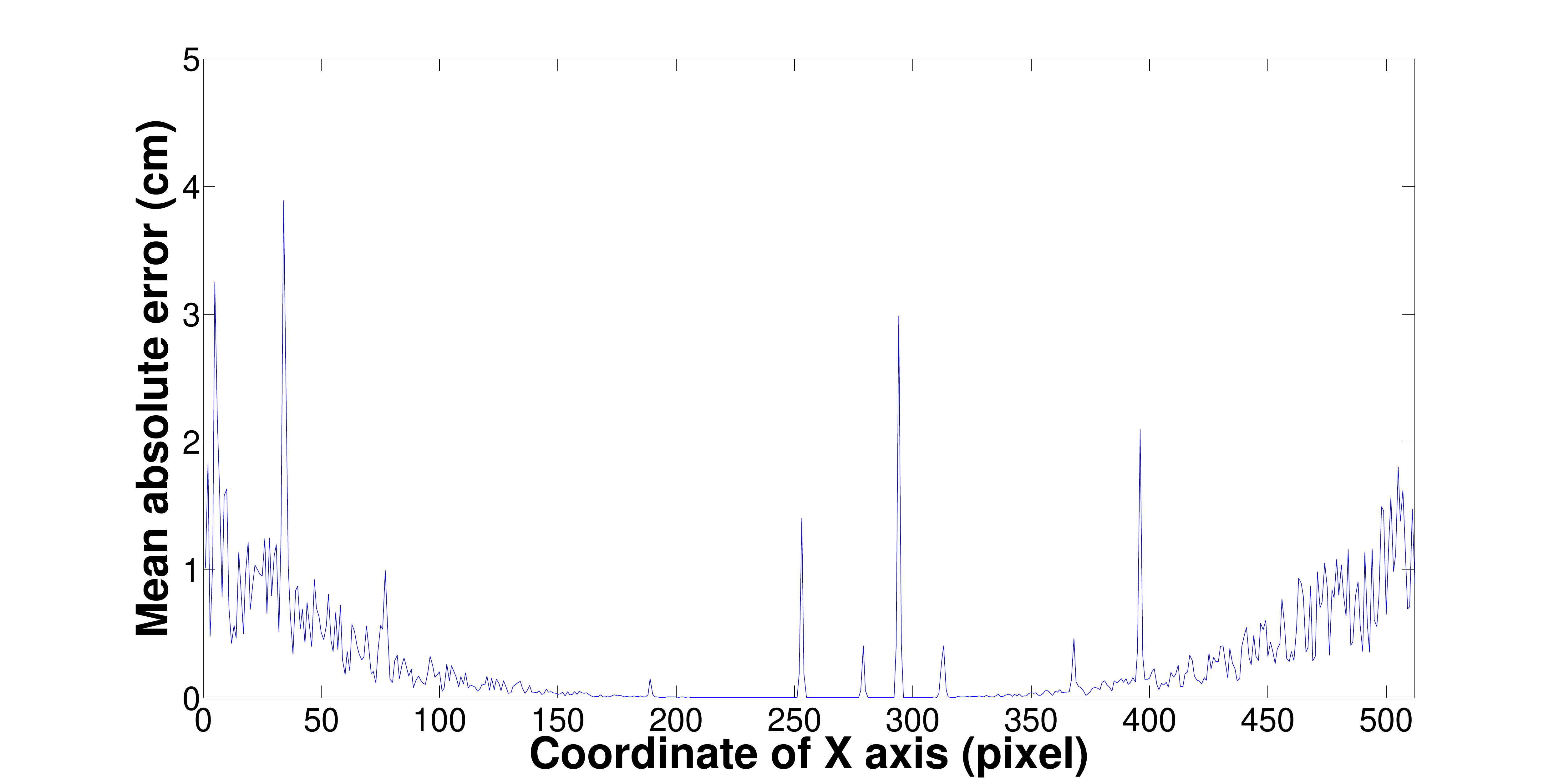}
	\par
	
	\protect\caption{\label{fig:distortion}Mean absolution error of the  pixels located in the middle row of  depth image.}
\end{figure}

The reason that calibration accuracy is affected by the mentioned angle is the distortion of the depth image. As shown in Fig. \ref{fig:calibration_reuslt}, the calibration error increases when the calibration wand leaves the center of Kinect's field-of-view. A further experiment to test the sensor's distortion is conducted where a flat plain is measured 2 m away from the Kinect v2, and 1000 frames of the depth image are obtained. The mean absolute errors of every pixels in the depth map are calculated. For the distortion is centrally symmetrical, pixels located in the middle row of the map are extracted and shown in Fig. \ref{fig:distortion}. It is shown that the mean absolute error on the dual edges of the depth image (pixels in x-coordinate from 0 - 150 and 350-520) is much higher than that in the middle of the depth image, leading that the recommended angle between the wand to the central line of Kinect is less than $26\degree$. Here, the big peaks in this figure is due to the noise in depth image. 

Based on the above analysis, within the recommended measurement range (i.e., distance: between 0.8 m and 3 m, angle: less than $26\degree$), 50 sets of calibration samples are collected and analyzed: the mean absolute error of calibration results is $0.73 \pm 0.85$ cm.

\subsection{Occlusion Compensation Verification}

To evaluate the effect of occlusion compensation component on the accuracy of occluded joints' measurement, the trajectories of joints with and without applying the component are compared together with the ground truth (Optitrack system).

\begin{figure}[!t]
	\centering
	\subfloat[]{
		\centering\label{fig:occlusionresult1}
		\includegraphics[width=9.5cm]{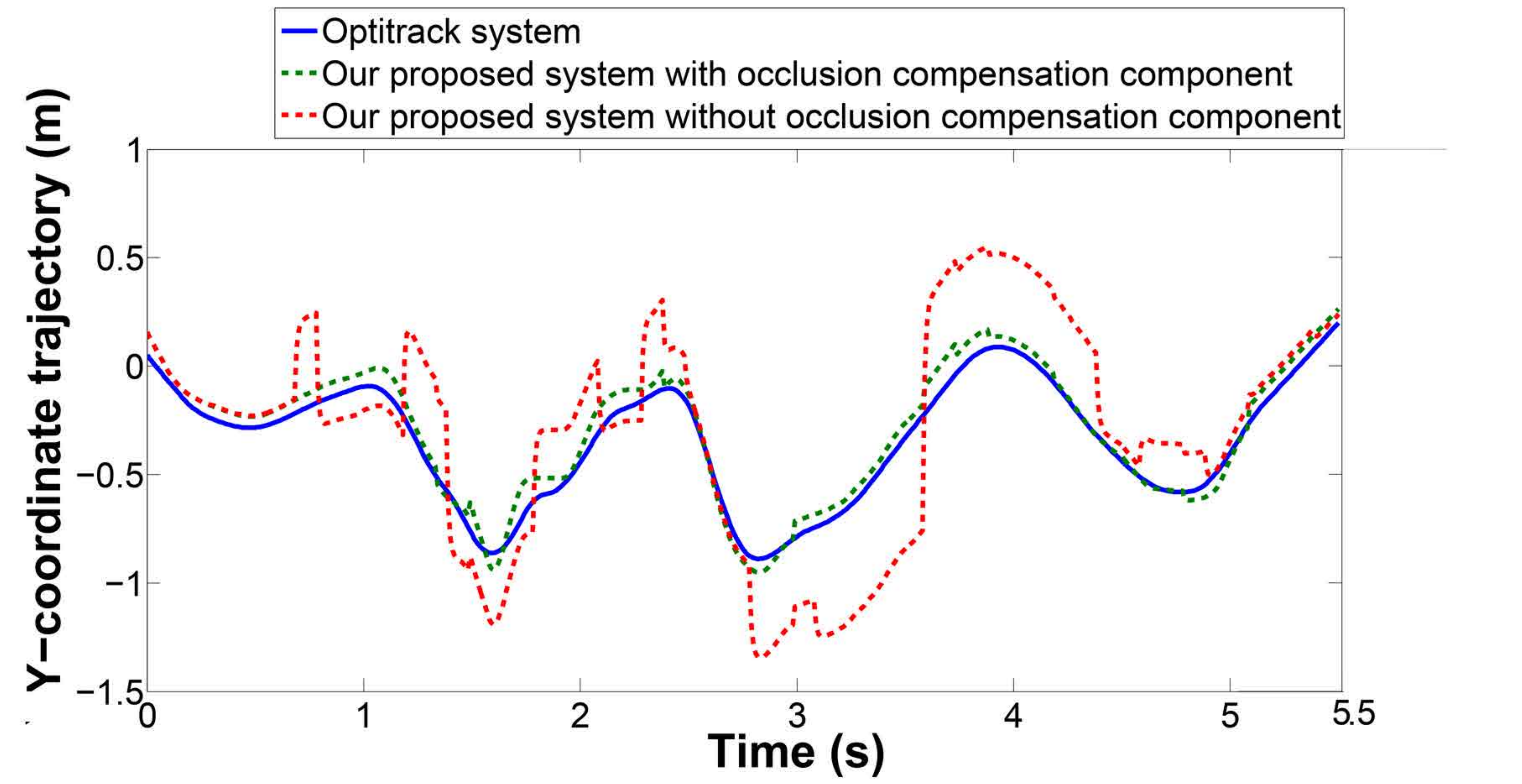}
		\par
	}
	\par
	
	\centering
	\subfloat[]{
		\centering\label{fig:occlusionresult2}
		\includegraphics[width=9.5cm]{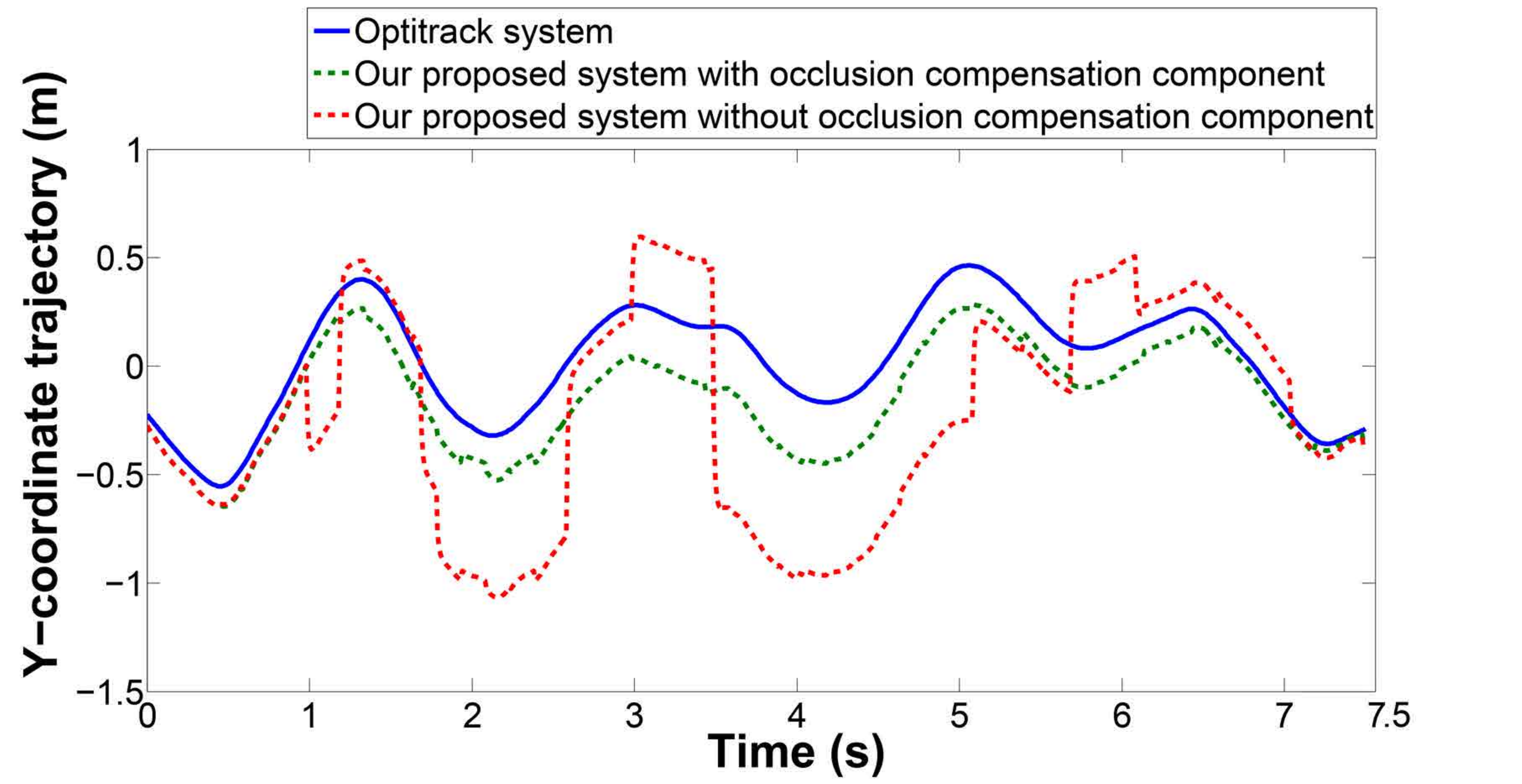}
		\par
	}
	\par
	\protect\caption{\label{fig:occlusionresult} 
	The y-coordinate trajectory of the joint of
	(a) elbow in Experiment 1 and (b) ankle in Experiment 2.
	The blue line indicates the ground-truth measured by Optitrack system. The green and red dots show the tracking result with and without applying the occlusion compensation component, respectively.
	}  
\end{figure}

In the experiment, two experiments are designed where 4 Kinects are set up in such way that part of them would obtain an occluded measurement. In Experiment 1, every participant swings his/her arms in a duration of 15 s. Here, the arms are partly occluded by the body and the shoulder, wrist, and elbow of the occluded arm are selected to be analyzed. In Experiment 2, every participant crosses his legs and make gait cycle in a duration of 15 s, where the hip, knee, ankle of both two legs are selected to be analyzed. The y-coordinate of the tracked joints (Elbow joint in Experiment 1 and Ankle joint in the Experiment 2) are shown in Fig. \ref{fig:occlusionresult}.

In Fig. \ref{fig:occlusionresult1}, the arm swings 2 times in a duration of 6 seconds where it is occluded by the body during the 2.5 s to 5.0 s in the field of view of one of the Kinects. It is shown that the result from the system without applying the occlusion compensation component is affected by the occluded measurements because the Kinect still returns its inferred result when it is not able to track the joint directly. With applying the occlusion compensation component, our proposed system can exclude the inferred result from occluded Kinect if there is an direct tracking from other Kinect sensors. Thus, the tracking results with occlusion compensation component (green dot curve) fit better with the Optitrack's ground-truth (blue curve) compared with that without collusion compensation component (red dot curve) in the time period between 2.5 s and 5.0 s.

In Fig. \ref{fig:occlusionresult2}, the participant steps forward and backward for 4 times in a duration of 7 s where one of the leg is occluded  in the middle 2 cycles from 3 s to 6.5 s. Similarly, the tracking results with applying occlusion compensation component (green dot curve) is better than that without occlusion compensation component (red dot curve). Whereas, the overall tracking results for both cases are not as good as the ones in Experiment 1 because the body tracking algorithm of Kinect has frequent mistakes on distinguishing the body posture for closing knees.

50 sets of occluded measurement samples are collected in the experiment 1 and 2, respectively. Compared with the ground-truth,  the mean absolute errors with occlusion compensation component are 13.3 cm and 19.4 cm for arm joints (shoulder, elbow, and wrist) and leg joints (hip, knee, and ankle) compared with that without the occlusion compensation component which are 18.5 cm and 23.2 cm, respectively (Table \ref{table:occlusionresult}). 

\begin{table}
	\protect\caption{\label{table:occlusionresult}Tracking Error with and without the occlusion compensation component}
	\begin{tabular}{|>{\centering}p{2cm}|>{\centering}p{2.5cm}|>{\centering}p{2.5cm}|}
		\hline
		\multirow{2}{*}{Tracked Joints}&
		\multicolumn{2}{c|}{Mean Absolute Error (cm)} 
		\tabularnewline\cline{2-3}
		
		&
		With Occlusion Compensation Component &
		Without Occlusion Compensation Componen
		\tabularnewline\hline

		Arm Joints (shoulder, elbow, and wrist)&
		13.3&
		18.5
		\tabularnewline\hline
		
		Leg Joints (hip, knee, and ankle) &
		19.4&
		23.2
		\tabularnewline\hline        
		
	\end{tabular}
\end{table}

\subsection{Accuracy Comparison between Trilateration Systems}

In this experiment, the Optitrack system and Kinect clients run simultaneously to measure the movements of body parts. Four joints of the human upper body are selected (i.e., head, shoulders, elbows and wrists) for tracking. Each participant completes 4 cycles of gaits in the duration of 35 s. To quantitatively evaluate the overall performance of proposed system, three trilateration systems (linear trilateration system \cite{SO_1}, geometric trilateration system \cite{SO_2} and our proposed nonlinear trilateration system) are applied to process the measurement data obtained from the Kinect sensors, based on which, the overall accuracy, the accuracy for different body parts are analyzed (shown as follows). Here, for different coordinate settings in the trilateration systems, the analysis is based on the relative movement of joints in the same time period.

\begin{figure}[!t]
	\centering
	\includegraphics[width=9.5cm]{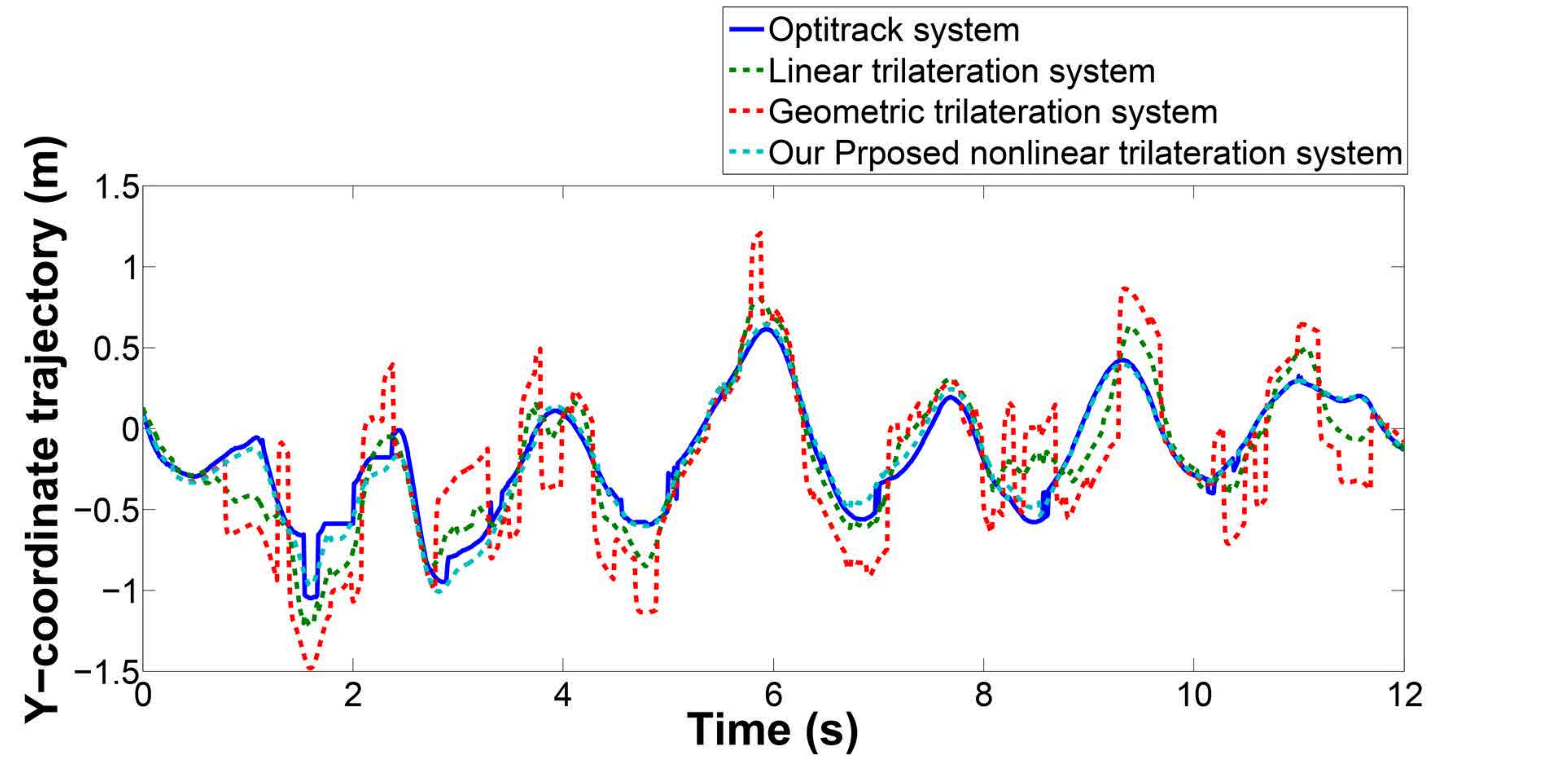}
	\par
	
	\protect\caption{\label{fig:comparison_1} The y-coordinates of the trajectories  of 
		right wrist, where blue, green, red, and cyan curves indicate its movement measured by the Optitrack system, linear trilateration system, geometric trilateration system, and our proposed nonlinear trilateration system, respectively. }
\end{figure}

Fig. \ref{fig:comparison_1} shows the tracking results of one participant who conducts 6 gait cycles with arms swing, meaning that his right wrist makes periodic movements. Compared the results from the mentioned four trilateration systems, the large deviation (maximum of 5 cm) occurs in the extreme swinging positions  while it is less than 1 cm in other gait phases for all of the three trilateration systems. The reason is that the Kinect sensor has big measurement error when the tracked joint has abrupt velocity variance. Our proposed nonlinear trilateration system has the smallest tracking error  compared against that of the other trilateration systems due to the robustness feature of its nonlinear optimization solution.

\begin{table}
	\protect\caption{\label{table:systemcomparison}Tracking accuracy comparision between the linear trilateration system, geometric trilateration system, and our proposed nonlinear trilateration system}
	\begin{tabular}{|>{\centering}p{1.5cm}|>{\centering}p{1.8cm}|>{\centering}p{1.8cm}|>{\centering}p{1.8cm}|}
		\hline
		\multirow{2}{*}{Tracked Joints}&
		\multicolumn{3}{c|}{Mean Absolute Error (cm)} 
		\tabularnewline\cline{2-4}
		
		&
		Linear Trilateration System \cite{SO_1} &
		Geometric Trilateration System \cite{SO_2} &
		Our Proposed Nonlinear Trilateration System
		\tabularnewline\hline

		Head&
		5.3&
		5.5&
		5.3
		\tabularnewline\hline
		
		Shoulder &
		10.5&
		11.8&   
		7.6     
		\tabularnewline\hline        
		
		Elbow &
		8.6&
		13.9&  
		7.3      
		\tabularnewline\hline    
		
		Wrist &
		18.3&
		25.0&   
		14.2    
		\tabularnewline\hline    
		
		Average &
		10.8&
		14.1&
		8.7        
		\tabularnewline\hline            
	\end{tabular}
\end{table}

 By comparing the mean absolute errors of the tracked joints in the mentioned different motion capture systems (shown in Table \ref{table:systemcomparison}), accuracy evaluation is quantitatively provided. The maximum error occurs in the measurement of wrists which has the longest movement trajectory and biggest velocity variance. One big factor that the accuracy of Kinect sensor is much affected by velocity is the low frame rate of Kinect's depth sensor which is 30 Hz at most. On the other hand, Kinect's resolution also has a limitation on measuring the end-side joints (e.g. wrists and ankles) which correspond with fewer pixels in the depth image.

Furthermore, the three trilateration systems have similar accuracy for tracking head and big difference in accuracy for the other three joints (shoulder, elbow, and wrist). For the reason of similar accuracy, Kinect sensor is able to obtain accurate head tracking due to the fact that the head's moving is relatively stable and hardly occluded. The lowest accuracy of the geometric trilateration system is due to its restriction of sensor number, meaning only geometric trilateration system applies 3 Kinect sensors whereas the other systems apply 4 Kinect sensors. Due to the feature of the nonlinear least square optimization approach, our proposed nonlinear trilateration system has the highest accuracy and robustness \cite{trilateration_1}. Specifically, our proposed system has a higher average accuracy by $24.1\%$ and $38.3\%$ compared with the linear and geometric trilateration systems, respectively.

\subsection{Efficiency Analysis}

An experiment is conducted to evaluate the system's performance on real-time processing and clock synchronization. Here, our proposed system consists of 1 server and 4 clients where the server and client time of sending and receiving messages and  completing specific functions (such as server's trilateration computation, occlusion compensation, etc. and client's human joint observation) are recorded, separately. Thus, the real-time processing character of the proposed system can be evaluated by calculating the time cost starting from obtaining the depth image on the client side to completing the trilateration computation on the server side; the clock synchronization can be tested by comparing the difference between the local time in clients and the synchronized time on the server side. Totally, 18000 frames (i.e., in 600 s in fps of 30) of data are recorded in the experiment. The statistical results of the synchronization results are shown in Table \ref{table:synchronizationevaluation}.
\begin{table}
	\protect\caption{\label{table:synchronizationevaluation}The statistics of  processing time cost and the clock error}
\centering	\begin{threeparttable}
\begin{tabular}{|>{\centering}p{2cm}|>{\centering}p{2cm}|>{\centering}p{2cm}|}
		\hline
		Time Range (ms)&
		Frame Statistics Based on Processing Time Cost  &
		Frame Statistics Based on Clock Error \tnote{*}
		\tabularnewline\hline

		[33,$+\infty$)&
		7&
		0
		\tabularnewline\hline

		[22,33)&
		11926&
		0
		\tabularnewline\hline

		[11,22)&
		6064&
		0
		\tabularnewline\hline
		
		[4,11)&
		0&
		0
		\tabularnewline\hline
		
		[2,4)&
		0&
		258
		\tabularnewline\hline
		
		(0,2)&
		0&
		71740		
		\tabularnewline\hline

	\end{tabular}
	
	\begin{tablenotes}
		\item[*] The clock errors are recorded from 4 clients, leading that there are totally $18000\times4=72000$ sets of clock error samples.
	\end{tablenotes}
	\end{threeparttable}
\end{table}

The frames excluded by synchronization component due to the error in the network transmission (3 frames) and those whose processing time is more than $\frac{1000}{30}\approx33$  ms (7 frame) would affect the real-time processing feature (Table \ref{table:synchronizationevaluation}), Thus, our proposed system guarantees the real-time feature in $99.94\%$ (17990 frames in 18000 frames) of its measurement process. It is shown that all clients dynamically synchronize their clocks with the server leading that all clock errors are less than 4 ms. More specifically, $99.36\%$ of the clock errors are less than 2 ms.

\section{Conclusion}

In our proposed system, the calibration component registers relative positions of sensors and transfer coordinates for the further calculation. After that, the occlusion compensation component filters the inferred measurements coming from Kinect v2 sensors for the trilateration component which combines the measurements from multiple sensors to obtain the optimized human body tracking. For all times, the synchronization component coordinates between the clients and server, which enables the real-time feature of the system. The aforementioned 4 components are quantitatively evaluated in the experiment, verifying the validity of the proposed motion capture system.


%





\ifCLASSOPTIONcaptionsoff
  \newpage
\fi



%

\bibliographystyle{ieeetr}
\bibliography{Bibliov}


%

\begin{IEEEbiography}[{\includegraphics[width=1in,height=1.25in,clip,keepaspectratio]{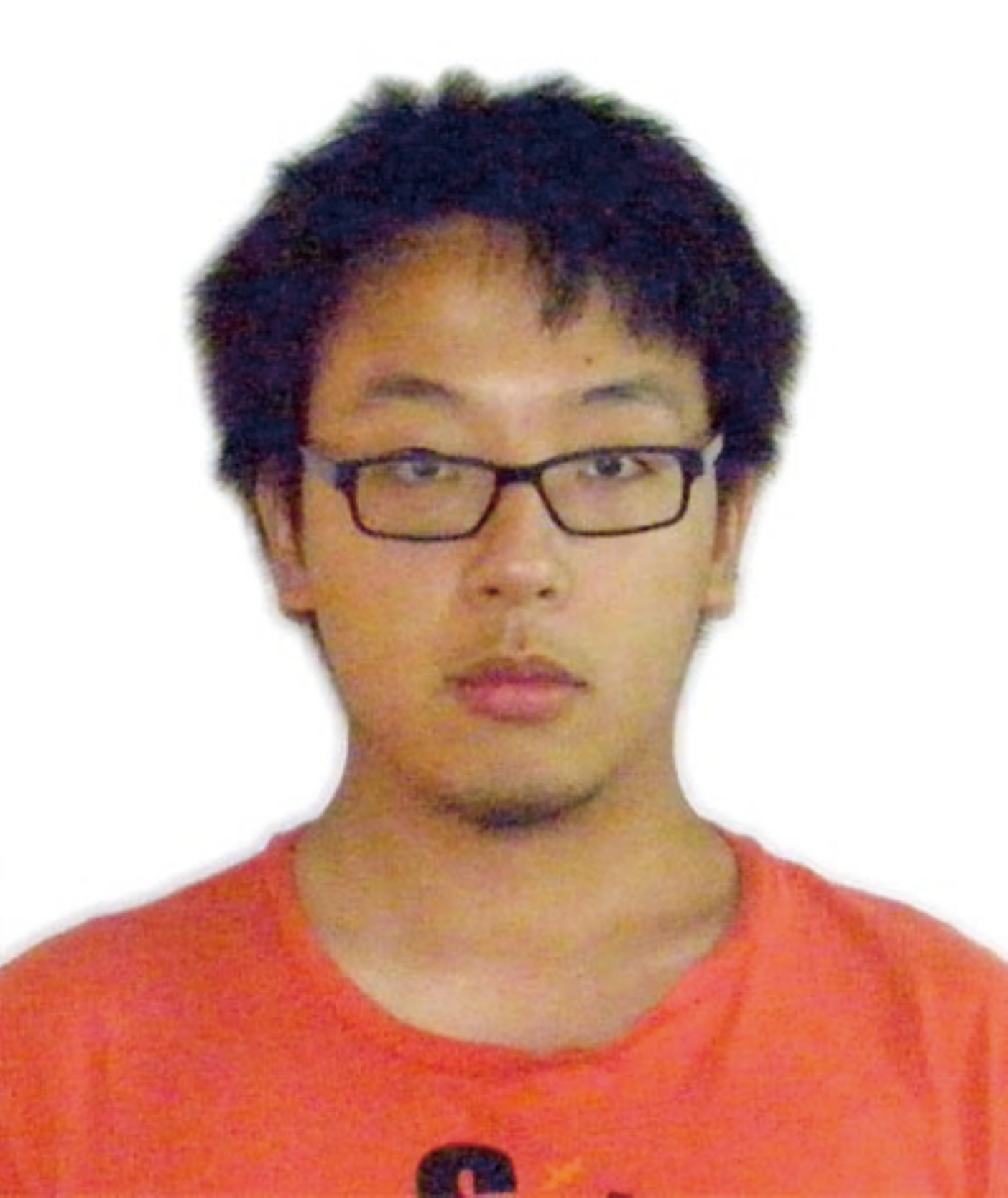}}]{Bowen Yang}
	received his B.Eng. in Computer Science from Central South University in 2014 and started his M.Sc. at the University of Ottawa in the same year. He is currently working on the project of 3D Sensing and Tracking of Human Gait Movement, which involves finding an optimal solution to improve the accuracy of multiple Microsoft Kinect v2 sensors for medical purposes. His research interests include computer vision, nonlinear optimization, and calibration strategy.
\end{IEEEbiography}


\begin{IEEEbiography}[{\includegraphics[width=1in,height=1.25in,clip,keepaspectratio]{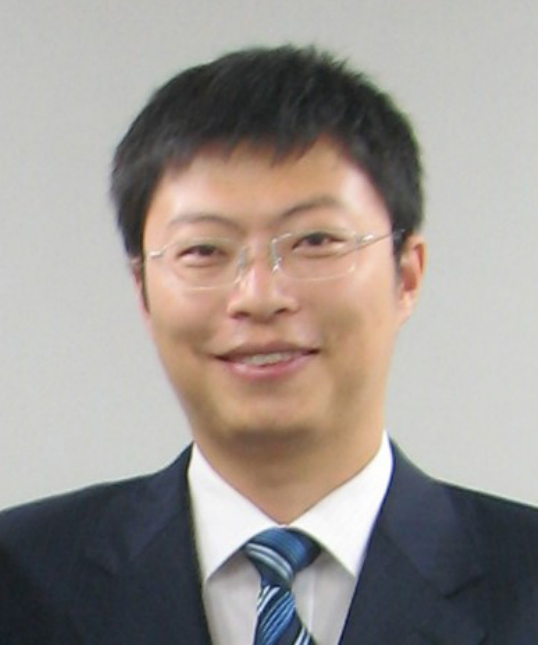}}]{Haiwei Dong (M’12-SM’16)}
received his Dr.Eng. in Computer Science and Systems Engineering and his M.Eng. in Control Theory and Control Engineering from Kobe University (Japan) and Shanghai Jiao Tong University (P.R.China) in 2010 and 2008, respectively. He is currently with the University of Ottawa. Prior to that, he was appointed as a Postdoctoral Fellow at New York University, a Research Associate at the University of Toronto, a Research Fellow (PD) at the Japan Society for the Promotion of Science (JSPS), a Science Technology Researcher at Kobe University, and a Science Promotion Researcher at the Kobe Biotechnology Research and Human Resource Development Center. His research interests include robotics, haptics, control and multimedia. He is a Senior Member of the IEEE.
\end{IEEEbiography}


\begin{IEEEbiography}[{\includegraphics[width=1in,height=1.25in,clip,keepaspectratio]{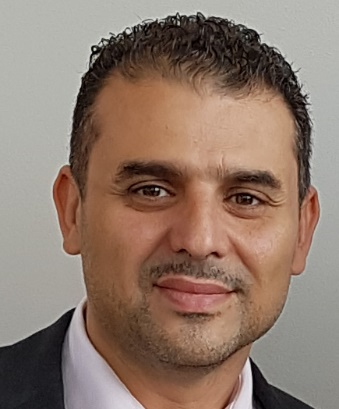}}]{Abdulmotaleb El Saddik (M’01-SM’04-F’09)}
	is Distinguished University Professor and University Research Chair in the School of Electrical Engineering and Computer Science at the University of Ottawa. His research focus is on multimodal interactions with sensory information in smart cities. He has authored and co-authored four books and more than 550 publications and chaired more than 40 conferences and workshop. He has received research grants and contracts totaling more than \$18 M. He has supervised more than 120 researchers and received several international awards, among others, are ACM Distinguished Scientist, Fellow of the Engineering Institute of Canada, Fellow of the Canadian Academy of Engineers and Fellow of IEEE, IEEE I\&M Technical Achievement Award and IEEE Canada Computer Medal.
\end{IEEEbiography}
\vfill




\end{document}